\documentclass{article}

\usepackage{arxiv}

\usepackage[utf8]{inputenc}
\usepackage[T1]{fontenc} 
\usepackage{hyperref}       
\usepackage{url}            
\usepackage{booktabs}      
\usepackage{amsfonts}       
\usepackage{nicefrac}       
\usepackage{microtype}      
\usepackage{graphicx}
\usepackage{natbib}
\usepackage{doi}
\usepackage{hyperref}
\usepackage{wrapfig}
\usepackage{multirow}
\usepackage{geometry}
\usepackage{amsmath}
\DeclareUnicodeCharacter{2212}{-}
\usepackage{graphicx}
\usepackage{subfigure}
\usepackage{rotating}
\usepackage{wrapfig}
\usepackage{listings}
\usepackage{caption}
\usepackage{subcaption}
\title{WinoWhat: A Parallel Corpus of Paraphrased WinoGrande Sentences with Common Sense Categorization}
\date{} 			

\author{ \href{https://orcid.org/0000-0003-1565-4077}{\includegraphics[scale=0.06]{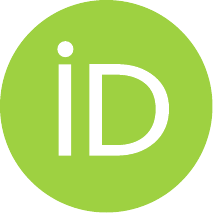}\hspace{1mm}Ine Gevers} \\
	CLiPS\\
	University of Antwerp\\
	Antwerp, 2000 \\
	\texttt{ine.gevers@uantwerpen.be} \\
	\And
    \href{https://orcid.org/0009-0005-8179-888X}{\includegraphics[scale=0.06]{orcid.pdf}\hspace{1mm}Victor De Marez} \\
	CLiPS\\
	University of Antwerp\\
	Antwerp, 2000 \\
	\texttt{victor.demarez@uantwerpen.be} \\
    \And
    \href{https://orcid.org/0000-0002-5288-1650}{\includegraphics[scale=0.06]{orcid.pdf}\hspace{1mm}Luna De Bruyne} \\
	CLiPS\\
	University of Antwerp\\
	Antwerp, 2000 \\
	\texttt{luna.debruyne@uantwerpen.be} \\
    \And
	\href{https://orcid.org/0000-0002-9832-7890}{\includegraphics[scale=0.06]{orcid.pdf}\hspace{1mm}Walter Daelemans} \\
	CLiPS\\
	University of Antwerp\\
	Antwerp, 2000 \\
	\texttt{walter.daelemans@uantwerpen.be} \\
}

\renewcommand{\shorttitle}{\textit{arXiv} Template}

\hypersetup{
pdftitle={WinoWhat: A Parallel Corpus of Paraphrased WinoGrande Sentences with Common Sense Categorization},
pdfsubject={q-bio.NC, q-bio.QM},
pdfauthor={Ine Gevers, Victor De Marez, Luna De Bruyne, Walter Daelemans},
pdfkeywords={Winograd Schema Challenge, parallel paraphrased corpus, common sense categorization, benchmark memorization},
}

\begin{document}
\maketitle

\begin{abstract}
In this study, we take a closer look at how Winograd schema challenges can be used to evaluate common sense reasoning in LLMs. Specifically, we evaluate generative models of different sizes on the popular WinoGrande benchmark. We release WinoWhat, a new corpus, in which each instance of the WinoGrande validation set is paraphrased. Additionally, we evaluate the performance on the challenge across five common sense knowledge categories, giving more fine-grained insights on what types of knowledge are more challenging for LLMs. Surprisingly, all models perform significantly worse on WinoWhat, implying that LLM reasoning capabilities are overestimated on WinoGrande. To verify whether this is an effect of benchmark memorization, we match benchmark instances to LLM trainingdata and create two test-suites. We observe that memorization has a minimal effect on model performance on WinoGrande.
\end{abstract}

\section{Introduction}
\begin{wrapfigure}[15]{r}{0.5\textwidth}
    \vspace{-20pt}
    \centering
    \includegraphics[width=0.5\textwidth]{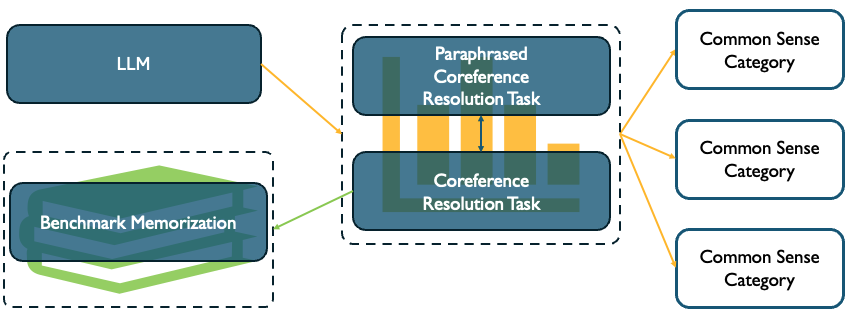}
    \caption{Illustration of the workflow in this study. We evaluate LLMs on WinoGrande, and on its paraphrased variant. We further compare performance per common sense knowledge category, and check for benchmark memorization.}
    \label{fig:workflow}
\end{wrapfigure}

While including common sense knowledge in NLP-systems has been a longstanding goal, evaluating this proves to be a non-trivial task. 
From early on, research used coreference resolution tasks to measure world knowledge and reasoning abilities in machine learning systems. In 2011, the Winograd Schema Challenge was developed, a small test set of 273 instances in which a pronoun has to be disambiguated given two possible antecedents in a short text \citep{levesque2012winograd}.
Where early models failed, transformer-based models quickly achieved remarkable performance on this test. However, researchers objected that this does not prove that models have or use common sense; rather, they could rely on superficial patterns and dataset artifacts \citep{kocijan2023defeat}. Therefore, a large adversarial benchmark was created: WinoGrande \citep{sakaguchi2021winogrande}. Here, the challenge is to decide which of two options is the correct one in a fill-in-the-blank token `\_'. This benchmark is frequently used in combination with other benchmarks to evaluate the performance of new LLMs on common sense reasoning.\\
In this study, we evaluate various open-source model families -- Gemma 2 \citep{team2024gemma}, LlaMA 2 \citep{touvron2023llama}, and OPT \citep{zhang2022opt} -- on WinoGrande. An overview of the workflow in this study can be found in Figure \ref{fig:workflow}. 
We present a new parallel corpus of the WinoGrande validation set: WinoWhat, in which we paraphrase each sentence so the `\_' token is at the end of the sentence. This transformation makes the task more natural for decoder-only methods and at the same time allows to test whether the performance of LLMs on WinoGrande is robust against paraphrasing (\textbf{RQ1}).\\ 
While existing works mainly evaluate models on the benchmark as a whole, we define common sense knowledge categories that are crucial to disambiguate the sentence, and evaluate models on each category separately. This allows us to investigate our second research question (\textbf{RQ2}): What types of common sense knowledge are more challenging for LLMs? Such an analysis provides insights into the more fine-grained strengths and weaknesses of ML systems on common sense reasoning tasks. Instead of creating new benchmarks to focus on one category of common sense knowledge, we suggest using one task setup, which allows us to compare results on different categories without added noise (e.g., different answer formats, different dataset artifacts, etc.).\\
To verify to what extent data leakage plays a role in LLMs' performance on WinoGrande, we check how many instances of the validation set are included in the pre-training data of LLMs. Further, we create two test-suites; one from which we know that it has been included in LLM pre-training data (i.e., the Winograd Schema Challenge), and one from which we can assume that it has not been seen (i.e., the WinoGrande test set). Comparably to RQ1, we paraphrase each. This answers \textbf{RQ3}: What is the role of data memorization in the performance of LLMs on coreference resolution tasks? \\
The rest of the paper is structured as follows: in Section \ref{related works}, we summarize relevant literature about disambiguation tasks, common sense categorization, and benchmark memorization. In Section \ref{methodology}, we present in more detail the data, models, evaluation metrics, and the creation of WinoWhat. Further, in Section \ref{results}, we present the results of our experiments, and the final Section \ref{conclusion} concludes our research, giving an overview of the findings and suggestions for further research.

\section{Related Work} \label{related works}
\subsection{Coreference resolution and common sense reasoning}
Incorporating common sense knowledge into machine learning methods has been a goal since its very beginning (e.g., \citealp{mccarthy1959programs}). However, given the increasing black-box nature of LLMs, it is hard to evaluate whether models have/use common sense knowledge. From early on, sentence disambiguation tasks have been suggested to measure the ability of models to employ common sense knowledge: the assumption being that syntax alone is not enough for the model, and common sense knowledge is needed to determine for instance which noun a pronoun refers to \citep{BROWNING2023104031}. 
An important concept here is bridging, for which the model needs to make inferences about relationships between entities in the world that are not explicitly mentioned in the text \citep{kobayashi2020bridging,hou2018unrestricted}. 
Since sentence disambiguation and coreference resolution tasks are presented as a proxy to evaluate common sense knowledge, over the years different approaches have been suggested to teach models common sense in order to improve performance on these tasks. In the early 2000s, most coreference resolvers did not include external knowledge sources, relying instead on morpho-syntactic features. The development of large-scale knowledge bases, which were used as features in a baseline resolver, improved results \citep{rahman2011coreference}. 
Then, with the advent of larger models and more training data, transformer models also relied on external knowledge bases which are generally stored in triplets \citep{liu2023brief}. 

\subsection{The Winograd Schema Challenge}
A popular coreference task is the Winograd Schema Challenge (WSC) \citep{levesque2012winograd}. 
Based on the work of \citet{winograd1972understanding}, the challenge uses `schemas' -- pairs of twin sentences whose intended meaning can be flipped by changing one word (the `special word') -- to probe ML-methods' ability to reason over natural language\footnote{A classic example is `The trophy didn't fit in the brown suitcase because \textit{it}'s too [small/big].'}. The schemas have three criteria: (1) simple to solve for humans; (2) not solvable by selectional restrictions (i.e., no statistical advantage for one option); (3) google-proof. 
Over time, newer versions of the challenge were released, which were made in the same format. These datasets are either created by human annotations, or generated by LLMs. For instance, \citet{zahraei-emami-2024-wsc} use Tree-of-Experts to generate new WSC instances, presenting 3,026 LLM-generated sentences.
Similarly, \citet{sun-emami-2024-evograd} present EvoGrad, a hybrid method to generate new adversarial WSC instances that feature minor alterations and rewordings using human annotations, ChatGPT, and WordNet.
Since WSC and related benchmarks are in English, the challenge was also translated in other languages such as German, French, and Russian \citep{emelin2021wino}.
The task has also been reformulated to evaluate implicit biases in LLMs, with resulting benchmarks such as WinoGender and WinoBias \citep{rudinger-etal-2018-gender,zhao-etal-2018-gender}.\\
By 2019, large pre-trained transformer models were reported to achieve over 90\% accuracy on WSC \citep{kocijan2023defeat}. 
Whereas the initial hypothesis was that systems would need common sense to solve the WSC, there is no proof that this is the case. Indeed, LLMs can rely on superficial pattern recognition and data memorization to solve the task, leading to the conclusion that these results are not indicative of common sense acquisition \citep{sakaguchi2021winogrande}. 
Furthermore, questions are raised about the quality and implicit biases of WSC, such as lax evaluation, annotation artifacts, and knowledge leakage from training data \citep{kocijan2023defeat,elazar2021back}. 
\citet{trichelair2018evaluation} also show that the `google-proof' condition, that stipulates that it should not be solvable via statistics learnt from large corpora associating one option to other components in the sentence, is not true for all instances in WSC. 
In an effort to address these limitations, adversarial variants of the WSC are presented. For instance, \citet{han2024concept} adapt the options so that they are more associated with the wrong answer, and \citet{trichelair2018reasonable} switch the position of the options in the texts where possible. Both report a decrease in model performance. \citet{abdou-etal-2020-sensitivity} show that models are not robust against linguistic perturbations such as changes in tense, gender, or synonym substitution in WSC sentences.
Additionally, the WinoGrande benchmark is introduced \citep{sakaguchi2021winogrande}. This benchmark is of a much larger scale (44K instances compared to the 273 in WSC), and employs an algorithm to reduce biases that machines can exploit to solve the task.

\subsection{Common sense knowledge categorization} \label{CS categorization}
To the best of our knowledge, research on WinoGrande discusses model results holistically (on the entire test or validation set), but we suggest connecting this to common sense knowledge categorization as an effective error analysis of the task. By measuring the performance per category, we can isolate reasoning deficiencies that are obscured by aggregated metrics.
There has been much effort on defining semantic categories to structure knowledge for NLP. \citet{SCHANK1972552} describes four main categories in their conceptual dependency theory: objects, actions, location, and time. \citet{jackendoff1992semantic} suggests common primitives such as entity, property, number, location, state, event, and activity. 
Other work only uses two high-level categories, such as social and physical \citep{sap2020commonsense}. Yet others define semantic categories within one common sense category; for instance, \citet{wang-etal-2021-semantic} include feelings and characteristics, interaction, and norms as sub-categories of social common sense. 
Additionally, different common sense categories are sometimes evaluated by specific independent benchmarks (e.g., spatial \citep{xu2017automatic,liu2022things}, temporal \citep{zhou2019going,aroca2021prost,hosokawa2024temporal,qin2021timedial}, numerical \citep{lin2020birds}, physical \citep{Bisk_Zellers_Le_bras_Gao_Choi_2020,storks2021tiered}, social \citep{sap2019socialiqa}, etc.). This can be problematic when comparing one model's ability to reason over various common sense categories, since each benchmark can have a different answer format (i.e., multiple choice, binary choice, open-ended) and structure. Other benchmarks that are more general, do not provide common sense categorizations. Therefore, we annotate the WinoGrande benchmark (a general-purpose benchmark) with which common sense knowledge is relevant when making the decision (i.e., what knowledge is needed when making the bridging inference). In a similar effort, \citet{zhang-etal-2020-winowhy} proposed 6 common sense categories to evaluate performance on the WSC: property, object, eventuality, spatial, quantity, and others.

\subsection{Benchmark memorization and contamination}
\citet{DBLP:journals/corr/abs-2406-04244} define benchmark data contamination (BDC) as LLM exposure to benchmark data during training, leading to inflated evaluation results. They outline contamination severities ranging from exposure to meta information about the benchmark or the task, to the benchmark data itself with labels. One main detection technique is $n$-gram overlap counting, as used by GPT-3 \citep{NEURIPS2020_1457c0d6} (13-gram) and GPT-4 \citep{achiam2023gpt} (40-gram). However, it requires full pre-training data access and can miss rephrasing \citep{DBLP:journals/corr/abs-2311-04850}. Additionally, \citet{wang2025generalization} find that factual or lexical tasks are particularly susceptible to memorization, while \citet{carlini2023quantifying} demonstrate that memorization increases with model size, data frequency, and sufficient context.\\
Since 2012, many WSC sentences have appeared in web text used to train LLMs \citep{elazar2021back}. RedPajama \cite{NEURIPS2024_d3449733} contains 58.2\% of WSC instances, while other datasets like The Pile  \citep{gao2020pile800gbdatasetdiverse} contain around 30\% \citep{elazar2024whats}. Such contamination inflates accuracy scores: \citet{emami-etal-2020-analysis} show significant accuracy drops when contamination is minimized.\\ 
In contrast, WinoGrande's creators mitigated contamination by keeping the test labels private. Regarding the validation set, only 1.1\% of this set appears online or in CommonCrawl between December 2020 and October 2023 \citep{li-etal-2024-open-source}, and the authors of GPT-4 self-report approximately 0.9\% contamination in a sample of 1,000 instances \citep{achiam2023gpt}. \citet{elazar2024whats} demonstrate that large pretraining corpora for LLMs did likely not encounter the WinoGrande test set, but they do not examine contamination of the validation set in these pretraining corpora. Thus, the precise effect of the contamination of the WinoGrande validation set is unknown, but for other benchmark data, it was previously shown that the effect of even minimal contamination can be underestimated \citep{singh2024evaluationdatacontaminationllms}. 

\section{Methodology} \label{methodology}

\subsection{Data} \label{methodology data}
In this study, we apply models on the WinoGrande benchmark, which was originally presented in 2019 as an adversarial dataset to the Winograd Schema Challenge (WSC) \citep{sakaguchi2021winogrande}.
Contrary to WSC, in which the sentence includes a pronoun that must be disambiguated given two candidate antecedents, the WinoGrande benchmark works with a fill-in-the-blank token `\_' (see Figure \ref{example_paraphrasing}). Additionally, every instance does not necessarily have a twin sentence.
The original paper reports human accuracy of 94\%, and model accuracy of 79.1\%, which is considerably lower than on WSC (over 90\%). The labels of the test set are not publicly available, which has led to research reporting on the validation set (see e.g., \citet{li2021systematic,sun-emami-2024-evograd,elazar2021back}). 
For that reason, we will also report on the validation set. This split consists of 1,267 instances, with a balanced label distribution.
The WinoGrande benchmark is also frequently used to evaluate new LLMs\footnote{It is unclear whether they report on the validation or test set.}. Recent evaluations include Gemma 2 27B at 83.7\% \citep{team2024gemma}, LlaMA 2 models ranging from 69.2\% (7B) to 80.2\% (70B) \citep{touvron2023llama}, GPT-4 (few-shot) achieving 87.5\% \citep{achiam2023gpt}, and Pythia 12B scoring 66.6\% \citep{biderman2023pythia}. 

\subsection{Models}
We focus on recent open-source Large Language Models. Since model size is a known factor in model performance, we select model families that have different sizes available. Specifically, we select Gemma 2 (2B, 9B, and 27B) \citep{team2024gemma}; LlaMA 2 (7B, 13B, and 70B) \citep{touvron2023llama}, and OPT (1.3B, 6.7B, 13B, and 66B) \citep{zhang2022opt} to evaluate the effect of paraphrasing WinoGrande, and for the evaluation per common sense category. Further, to evaluate benchmark memorization, we include two other models because their pre-training data is publicly available, contrary to the previously mentioned models: Pythia (1B, 1.8B, 6.9B, and 12B) \citep{biderman2023pythia} and LlaMA 1 (7B, 13B, 30B, and 65B) \citep{touvron2023llama-1}.\\
To evaluate model performance, we use partial evaluation, which calculates the summed log-likelihood for the tokens after each option in the text, selecting the one with the highest score \citep{trinh2018simple}. We choose this metric for three reasons:\\
1. It is the evaluation metric used in the Language Model Evaluation Harness \citep{eval-harness}, which is the base of the Huggingface Open-LLM Leaderboard\footnote{WinoGrande was included in the V1 of the leaderboard: \url{https://huggingface.co/docs/leaderboards/en/open_llm_leaderboard/archive}};\\
2. Preliminary experiments show that it works better than prompting, and \citet{trinh2018simple} show that it works better than full evaluation;\\
3. It is easily generalizable to different open-source models.

\subsection{Paraphrased corpus} \label{methodology paraphrasing}
\begin{figure*}
\includegraphics[width=15cm]{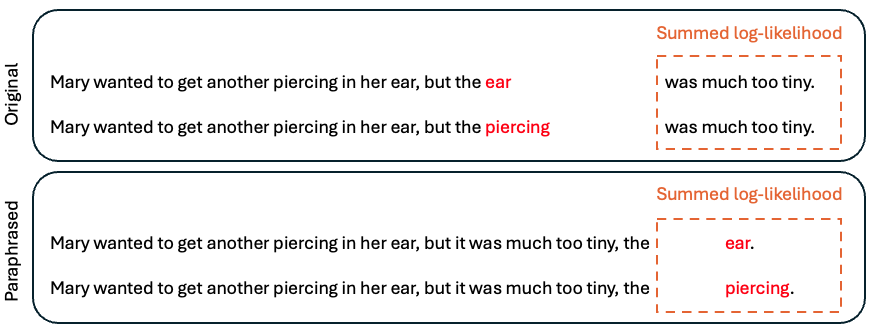}
\caption{An illustration of the paraphrasing and evaluation method. The option that is filled in the `\_'-token is in red. In the original example, the summed log-likelihood is calculated on the tokens after the option. In our paraphrased corpus, the option is at the end of the sentence, and the summed log-likelihood is calculated on the tokens inside the option.}
\label{example_paraphrasing}
\end{figure*}

To test the robustness of model performance on WinoGrande, we create WinoWhat: a parallel corpus in which we paraphrase the sentences. Additionally, this solves a limitation of the partial evaluation metric. Its main limitation is that it relies on the plausibility of the subsequent sequence, rather than directly measuring a model’s intrinsic token preference. This can conflate the model's understanding of the antecedent with grammatical or natural continuations. In contrast, with our paraphrased corpus, we position the target token at the end of the sequence, ensuring that the decision is based solely on the provided context. This allows for a more transparent evaluation of the model's ability to capture coreference and fill-in-the-blank cues. Contrary to the original partial evaluation that measures the summed log-likelihood on the tokens following the `\_' token, our method calculates it on the tokens of the options. An example is given in Figure \ref{example_paraphrasing}.\\ 
We prompt 5 SOTA LLMs (i.e., GPT-4o \citep{openai2024gpt4ocard}, OpenAI o1-preview \citep{openai2024openaio1card}, Gemini 2.0 Flash Thinking Experimental \citep{deepmind2024flashthinking}, Deepseek R1 \citep{deepseekai2025deepseekr1incentivizingreasoningcapability}, and LlaMA 3.2 90B Vision \citep{llama3.2}) to generate a paraphrased sentence given an input sentence, in which the `\_' token is at the end of the sentence. The generated options were manually checked, and the best option was selected for each sentence. However, in many cases (433), manual adjustments were still needed. The prompt for this task, and the distribution of which model's output is used, can be found in Appendix \ref{app:promptparaphrased}. 
In this stage, we also evaluate the validity of the sentences in the WinoGrande validation set. We notice that not all instances meet the requirements of WSC (e.g., not `google-proof', grammatical errors, etc.), which we remove in our paraphrased dataset. In total, we find 82 such cases\footnote{There are an additional 22 instances for which one annotator was not convinced of the quality. These instances were left out in the experiments, but for completeness are added in the released dataset.}.
On a sample of 100 paraphrased instances, we calculate the inter-annotator agreement between 3 annotators based on the following criteria: (1) Is the new sentence grammatical?; (2) Is the fill-in-the-blank token at the end of the sentence?\\
85\% of the texts are rated by all annotators as acceptable. 

\subsection{Common sense knowledge categorization} \label{methodology categorization}
\begin{wrapfigure}{r}{0.5\textwidth}
    \begin{center}
        \includegraphics[width=0.48\textwidth]{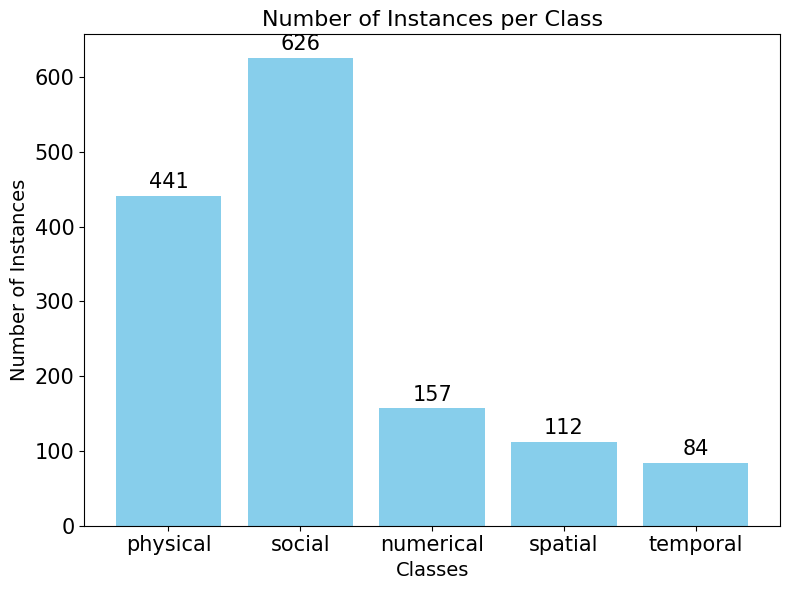} 
        \caption{Data distribution across common sense categories on the WinoGrande validation set.}
        \label{nr_instances_categories}
    \end{center}
\end{wrapfigure}
We categorized the coreference resolution instances according to the common sense knowledge type that is necessary to make the bridging inference. This categorization can function as data for an error analysis to detect what knowledge types are easier or harder for LLMs to solve.
Similarly to \citet{zhang-etal-2020-winowhy}, we select categories that have a broad coverage and are clearly distinguished from each other. We examine which categories are identified in existing benchmarks that evaluate common sense reasoning in NLP\footnote{e.g., see \url{https://cs.nyu.edu/~davise/Benchmarks/Text.html}}, which leads to five categories: physical, social, numerical, temporal, and spatial\footnote{Originally, we included causal as label, but removed this category: all instances in WinoGrande had this label, which was also noted by \citet{zhang-etal-2020-winowhy}.}.\\
We use LLMs to categorize the validation set. 
To identify the relevant common sense type, we prompt GPT-4o-mini to generate reasoning steps to solve the task. We then provide the input text and the generated reasoning steps to GPT-4o, which assigns one or more common sense categories to each instance. The prompts for these tasks are available in Appendix \ref{app:promptcategorization}. Annotation reliability is assessed by manually labeling 100 instances and comparing them with GPT-4o's labels, yielding a Kappa score of 0.64, which is a substantial agreement\footnote{The kappa scores per category: physical 0.63; social 0.68; numerical 0.58; temporal 0.72; spatial 0.59.} \citep{cohen1960coefficient}.
When applying our method on the entire validation set, we note a class imbalance; the physical and social categories are considerably larger than the other three, see Figure \ref{nr_instances_categories}.

\section{Results} \label{results}

\begin{table*}[ht!]
\centering
\begin{subtable}{}
\centering
\begin{tabular}{l|cc|cc|cc||cc|cc|cc}
\hline
\multicolumn{1}{c|}{} 
& \multicolumn{6}{c||}{\textbf{LlaMA 2}} 
& \multicolumn{6}{c}{\textbf{Gemma 2}} \\
\cline{2-13}
& \multicolumn{2}{c|}{7B} 
& \multicolumn{2}{c|}{13B} 
& \multicolumn{2}{c||}{70B} 
& \multicolumn{2}{c|}{2B} 
& \multicolumn{2}{c|}{9B} 
& \multicolumn{2}{c}{27B} \\
\hline
& \textbf{orig} & \textbf{transf} 
& \textbf{orig} & \textbf{transf} 
& \textbf{orig} & \textbf{transf} 
& \textbf{orig} & \textbf{transf} 
& \textbf{orig} & \textbf{transf} 
& \textbf{orig} & \textbf{transf} \\
\hline
\textbf{TOTAL}     & 0.69 & 0.58 & 0.72 & 0.62 & 0.78 & 0.70 
                   & 0.68 & 0.59 & 0.74 & 0.68    & 0.66 & 0.56 \\
\textbf{Physical}  & 0.71   & 0.61   & 0.73   & 0.63   & 0.77   & 0.73   
                   & 0.71   & 0.60   & 0.74   & 0.68    & 0.74   & 0.59   \\
\textbf{Social}    & 0.68   & 0.56   & 0.72   & 0.61   & 0.79   & 0.68   
                   & 0.68   & 0.57   & 0.73   & 0.67    & 0.60   & 0.54   \\
\textbf{Numerical} & 0.69   & 0.53   & 0.70   & 0.61   & 0.79   & 0.69   
                   & 0.63   & 0.62   & 0.75   & 0.62    & 0.69   & 0.51   \\
\textbf{Spatial}   & 0.71   & 0.61   & 0.76   & 0.65   & 0.75   & 0.70   
                   & 0.70   & 0.61   & 0.78   & 0.69    & 0.78   & 0.62   \\
\textbf{Temporal}  & 0.76   & 0.67   & 0.70   & 0.69   & 0.79   & 0.74   
                   & 0.65   & 0.62   & 0.74   & 0.71    & 0.67   & 0.54   \\
\hline
\end{tabular}
\caption{LlaMA 2 and Gemma 2 results on WinoGrande validation. The `orig' columns report the results on the original instances, the `transf' columns on the paraphrased instances.}
\label{tab:llama-gemma validation}
\end{subtable}
\vspace{0.5em}
\begin{subtable}{}
\centering
\begin{tabular}{l|cc|cc|cc|cc}
\hline
 & \multicolumn{2}{c|}{OPT 1.3B} 
 & \multicolumn{2}{c|}{OPT 6.7B} 
 & \multicolumn{2}{c|}{OPT 13B} 
 & \multicolumn{2}{c}{OPT 66B} \\
\hline
& \textbf{orig} & \textbf{transf} 
& \textbf{orig} & \textbf{transf} 
& \textbf{orig} & \textbf{transf} 
& \textbf{orig} & \textbf{transf} \\
\hline
\textbf{TOTAL}     & 0.60 & 0.53 & 0.66 & 0.54 & 0.65 & 0.56 & 0.69 & 0.58 \\
\textbf{Physical}  & 0.62   & 0.57   & 0.72   & 0.57   & 0.67   & 0.60   & 0.73   & 0.61   \\
\textbf{Social}    & 0.59   & 0.50   & 0.63   & 0.50   & 0.65   & 0.52   & 0.66   & 0.55   \\
\textbf{Numerical} & 0.57   & 0.49   & 0.62   & 0.58   & 0.63   & 0.54   & 0.68   & 0.57   \\
\textbf{Spatial}   & 0.56   & 0.61   & 0.65   & 0.61   & 0.63   & 0.61   & 0.67   & 0.61   \\
\textbf{Temporal}  & 0.50   & 0.55   & 0.57   & 0.58   & 0.61   & 0.53   & 0.66   & 0.57   \\
\hline
\end{tabular}
\caption{OPT results on WinoGrande validation. The `orig' columns report the results on the original instances, the `transf' columns on the paraphrased instances.}
\label{tab:opt validation}
\end{subtable}

\label{tab:winogrande-val}
\end{table*}

\subsection{Paraphrased corpus}
We report on the models' performance on WinoWhat. This allows us to compare the performance on the original texts to the paraphrased texts. If models truly generalize on the Winograd schemas, the performance should remain consistent; after all, the same information is conveyed, in the same task setup, only paraphrased. Additionally, we report on the performance per common sense category presented in Section \ref{CS categorization}. In Table \ref{tab:llama-gemma validation} and Table \ref{tab:opt validation}, subcolumn `orig' refers to the original texts in WinoGrande, `transf' to the paraphrased texts.\\
Considering the result on the WinoGrande validation set, we see that larger models generally perform better than their smaller variants, with LLaMA 2 70B performing the best. The error analysis comparing the performance of the same model across common sense categories shows that there is no one category that is impossible to be learned by a model, but there are fluctuations. Interestingly, we see that the category with the best results varies across model families: for LlaMA 2, there is no category that is consistently easier, while for Gemma 2 spatial is best, and for OPT physical. Temporal is consistently the worst category for OPT.\\
However, when comparing the original to the paraphrased task, we conclude that all models perform worse on the paraphrased corpus, and there is no common sense category that is robust against this transformation.\\
Our results challenge the assumption that LLMs apply reasoning when solving the WinoGrande task, suggesting they instead rely on dataset artifacts and/or memorization.
While \citet{sakaguchi2021winogrande} implemented an algorithm to automatically reduce machine-exploitable bias in their corpus, our results demonstrate that this might not be effective anymore in the LLM era.\\
We publicly release WinoWhat, consisting of the original WinoGrande validation set with the paraphrased counterparts and common sense categorizations\footnote{The dataset will be released later.}.

\subsection{Memorization}

\begin{table*}[ht!]
\hspace*{-0.8cm}%
\centering
\begin{minipage}[t]{0.50\textwidth}
\centering
\setlength{\tabcolsep}{3pt}
\begin{tabular}{l|cc|cc|cc}
\hline
\textbf{Model} & \multicolumn{2}{c|}{WG val} & \multicolumn{2}{c|}{WG test} & \multicolumn{2}{c}{WSC} \\
\cline{2-7}
               & orig & transf & orig & transf & orig & transf \\
\hline
LlaMA 2 7B    & 0.69 & 0.58   & 0.74 & 0.54   & 0.86 & 0.54 \\
LlaMA 2 13B   & 0.72 & 0.62   & 0.73 & 0.65   & 0.83 & 0.63 \\
LlaMA 2 70B   & 0.78 & 0.70   & 0.79 & 0.70   & 0.88 & 0.66 \\
\hline
Gemma 2 2B    & 0.68 & 0.59   & 0.73 & 0.61   & 0.83 & 0.64 \\
Gemma 2 9B    & 0.74 & 0.68   & 0.73 & 0.64   & 0.86 & 0.58 \\
Gemma 2 27B   & 0.66 & 0.56   & 0.58 & 0.57   & 0.76 & 0.51 \\
\hline
OPT 1.3B     & 0.60 & 0.53   & 0.58 & 0.50   & 0.72 & 0.54 \\
OPT 6.7B     & 0.66 & 0.54   & 0.52 & 0.56   & 0.82 & 0.56 \\
OPT 13B      & 0.65 & 0.56   & 0.68 & 0.56   & 0.81 & 0.56 \\
OPT 66B      & 0.69 & 0.58   & 0.71 & 0.52   & 0.82 & 0.58 \\
\hline
\end{tabular}
\caption{Accuracy results on WinoGrande (WG) validation, test and WSC for LlaMA 2, Gemma 2 and OPT.}
\label{tab:testsuites orig}
\end{minipage}
\hfill
\begin{minipage}[t]{0.50\textwidth}
\centering
\setlength{\tabcolsep}{3pt}
\begin{tabular}{l|cc|cc|cc}
\hline
\textbf{Model} & \multicolumn{2}{c|}{WG val} & \multicolumn{2}{c|}{WG test} & \multicolumn{2}{c}{WSC} \\
\cline{2-7}
               & orig & transf & orig & transf & orig & transf \\
\hline
LlaMA 1 7B    & 0.70 & 0.58   & 0.74 & 0.59   & 0.85 & 0.61 \\
LlaMA 1 13B   & 0.72 & 0.60   & 0.75 & 0.64   & 0.88 & 0.66 \\
LlaMA 1 30B   & 0.76 & 0.64   & 0.74 & 0.62   & 0.92 & 0.62 \\
LlaMA 1 65B   & 0.77 & 0.67   & 0.79 & 0.69   & 0.91 & 0.68 \\
\hline
Pythia 1B    & 0.54 & 0.53   & 0.57 & 0.54   & 0.71 & 0.50 \\
Pythia 2.8B  & 0.60 & 0.52   & 0.59 & 0.53   & 0.76 & 0.55 \\
Pythia 6.9B  & 0.61 & 0.52   & 0.58 & 0.56   & 0.77 & 0.52 \\
Pythia 12B   & 0.63 & 0.52   & 0.61 & 0.60   & 0.79 & 0.49 \\
\hline
\end{tabular}
\caption{Accuracy results on WinoGrande (WG) validation, test and WSC for LlaMA 1 and Pythia.}
\label{tab:testsuites llama1 pythia}
\end{minipage}
\hspace*{0.8cm}%
\end{table*}

Given the surprising drop in performance comparing  WinoGrande to WinoWhat, we investigate further what could cause this. 
While \citet{elazar2024whats} show that the test set of WinoGrande has probably not been seen by LLMs, this is not tested for the validation set. This is problematic, because research often reports on this split because of the absence of the test labels.
Therefore, it is crucial to verify how many instances of the WinoGrande validation set have been included in datasets used to pre-train LLMs. Specifically, we count how many instances appear entirely in the pre-training corpora.\\
Since the pre-training data for Gemma 2, LlaMA 2, and OPT models remains either undisclosed or inaccessible, we examine two LLMs with publicly available pre-training data: LlaMA 1 and Pythia, whose results are presented in Table \ref{tab:testsuites llama1 pythia}. These models were trained on RedPajama v1 \citep{together2023redpajama} and The Pile's training set \cite{gao2020pile} respectively.\footnote{Details about our method to check memorization can be found in Appendix \ref{app:memorization}.}\\
While we found that The Pile contains no contaminated instances, an interesting pattern emerges: as model size of Pythia increases, the performance gap between WinoGrande and WinoWhat widens, with WinoWhat accuracy remaining stable while WinoGrande scores improve (see column `WG val' in Table \ref{tab:testsuites llama1 pythia}).\\
An analysis of RedPajama v1 reveals 22 contaminated instances (1.7\% of the dataset), each appearing once and sourced from academic papers. To investigate potential memorization effects, we conduct a one-sided Mann-Whitney U test between performance on contaminated and non-contaminated instances across LlaMA 1 models (7B, 13B, 30B, and 65B). The results (see Table \ref{tab:contaminated-vs-winogrande} in Appendix \ref{app:memorization}), with $p$-values ranging from 0.054 to 0.267, show no significant evidence that LLaMA 1 models give preferential treatment to previously seen WinoGrande instances. However, similarly to Pythia, LlaMA 1 displays a consistent accuracy gap between WinoGrande and WinoWhat. Since this pattern is observed in all other models as well (Table \ref{tab:llama-gemma validation} and Table \ref{tab:opt validation}), it suggests that factors beyond simple memorization may be driving these performance differences.\\
To verify the role of contamination in later and more modern models with unknown pre-training data, we create two test-suites. 
Specifically, we take a sample ($n=100$) from the WSC dataset (of which we can assume that a substantial part has been memorized by LLMs \citep{elazar2024whats}), and paraphrase those; and we take a sample ($n=100$) from the test set of WinoGrande (of which we can assume that it has not been memorized by LLMs due to its private labels), which we label manually and paraphrase as well. 
We hypothesize that LLMs perform well on datasets that are polluted, but less so on unseen datasets. Therefore, we expect models to perform well on WSC, but below par on WSC paraphrased and WinoGrande test (both original and paraphrased). We summarize the results in Table \ref{tab:testsuites orig}. As expected, all models perform best on the original WSC benchmark. Paraphrasing almost always causes a drop in performance, regardless of the original source. 
The difference is biggest for the WSC benchmark, which is in line with our hypothesis given the pollution of this benchmark in LLMs' trainingdata. 
We still see a drop in performance for the WinoGrande test set, which is not included in the LLM trainingdata, when comparing the original sentences to the paraphrased ones. Together with our findings on Pythia and LLaMA 1, this indicates that there are other factors causing models to struggle with the paraphrased benchmark. We hypothesize that our evaluation metric better captures the model's performance on coreference resolution compared to the original partial evaluation (see Figure \ref{example_paraphrasing}), which could explain the drop in performance. Additionally, for larger and recent models, even though benchmark instances might not appear directly in the pre-training data, this does not exclude the possibility that it has been used during RLHF or instruction tuning, thereby compromising the validity of their performance on WinoGrande.

\section{Conclusion} \label{conclusion}
In this study, we take a closer look at how Winograd schema challenges can be used to evaluate common sense reasoning in LLMs. For this purpose, we focus on WinoGrande, a large adversarial benchmark created in 2019, frequently used to evaluate common sense in new LLMs. We select different generative model families, comparing models of the same family of different sizes. Specifically, we focus on Gemma 2, LlaMA 2, and OPT. To evaluate the models, we employ the partial evaluation metric. 
To address the limitations of the partial evaluation metric as outlined in Section \ref{methodology paraphrasing}, we create a parallel corpus to the WinoGrande validation set in which we paraphrase each text so the fill-in-the-blank token is at the end of the sentence (\textbf{RQ1}). 
In addition, we propose a new method to inspect performance on various common sense knowledge categories within the same task (\textbf{RQ2}). 
We select five categories: physical, social, numerical, spatial, and temporal. This approach can offer an in-depth error analysis, that sheds light on what types of knowledge are more challenging for LLMs. 
We publicly release WinoWhat, the parallel corpus to the WinoGrande validation set including the paraphrased sentences and the common sense categorization.
Our results show that while models perform well on the original WinoGrande validation set, they all perform worse on the paraphrased corpus, and all common sense categories are affected negatively. This questions the assumption that models apply reasoning, leaving the possibility for dataset artifacts or benchmark memorization.\\
To verify how much data memorization has an effect on the models' performance on the WinoGrande validation set (\textbf{RQ3}), we test whether instances that occur in pre-training data score significantly higher than instances that don't. 
We observe that the memorization of the validation set is minimal. Interestingly, we see that most contaminated instances come from academic publications citing examples from the benchmark. This again calls attention to the scraping methods to create large-scale pre-training data. 
Because the pre-training data of later models is unknown, we create two small ($n=100$) test-suites: one of which has been shown to be included in LLM training sets (i.e., the WSC benchmark) and one that is not seen by LLMs (i.e., the WinoGrande test set). We find that all models perform best on the WSC dataset, and paraphrasing causes a drop in performance. Since this is also the case for the WinoGrande test set, we conclude that there are other factors beside memorization that cause models to fail on the paraphrased task. 
Similarly to conclusions about the original Winograd Schema Challenge, this implies that we are again overestimating LLMs reasoning capabilities when using WinoGrande. Our new paraphrased corpus can be used to verify model generalization on the WinoGrande validation set.\\
In further research, we plan to inspect the information that is used by models to solve the task per common sense category using mechanistic interpretability: do models use similar information for each category? Do they rely on spurious correlations, and if so, which ones? 
Since data memorization does not seem to cause the drop in performance comparing the original to the paraphrased instances, we suggest to identify dataset artifacts that could be at the root of this. For instance, as previously done on WSC, do linguistic perturbations affect model performance?\\

\section*{Acknowledgments}
This research was made possible with a grant from the Fonds Wetenschappelijk Onderzoek (FWO) project 42/FA030100/9770, and funding from the Flemish Government under the “Onderzoeksprogramma Artificiële Intelligentie (AI) Vlaanderen” programme.

\bibliographystyle{unsrtnat}
\bibliography{references}

\appendix
\section{Limitations}
While, to the best of our knowledge, this is the first time the WinoGrande validation set has been annotated for common sense knowledge categories, this approach has possible shortcomings. First, the agreement between a human annotator and the labeling by GPT-4o shows a substantial agreement, but there will be cases with incorrect labeling. Therefore, we talk about aggregated results across categories in this study, since we're interested in trends, but for even more fine-grained interpretations this categorization should possibly be corrected manually. \\
Further, as is unfortunately still a trend in NLP-research, this dataset is in English, excluding lower-resource languages. Further research could translate our benchmark to other languages. \\
During the process of paraphrasing the original instances, we applied a strict quality check, which excluded 82 instances from the original dataset. While we believe this improves the quality of the resulted paraphrased dataset, this means we cannot make a perfectly aligned comparison to the original dataset.\\
Since we wanted to mitigate shortcomings of the partial evaluation metric, we paraphrased WinoGrande so the fill-in-the-blank token appears at the end of the sentence. While we argue that this setup is more natural for decoder-only models, and this evaluation metric is better suited to capture model performance on coreference resolution tasks rather than measuring natural continuations of the sentence, this resulted in a higher number of cleft-constructions. By adapting the evaluation method so it calculates the summed log-likelihoods on the tokens in the option rather than on the tokens after the option, this obscures whether the difference in performance is a result of the paraphrasing, or of the evaluation method. To verify this, we aim to construct a third level, in which we paraphrase without the constraint of putting the `\_'-token at the end of the sentence, allowing us to use the original partial evaluation method. This would indicate whether the drop in performance is caused by the paraphrasing itself, or by the evaluation metric.\\
Finally, our method of finding data contamination in pre-training data was on the data level only, not taking into account the semantic or information level \citep{DBLP:journals/corr/abs-2406-04244}. Methods such as ours relying on string matching methods might miss certain instances, such as rephrasings \citep{DBLP:journals/corr/abs-2406-04244}. Furthermore, such methods are only possible when access to pre-training corpora is public \citep{DBLP:journals/corr/abs-2311-04850}.

\section{Generating paraphrased sentences}\label{app:promptparaphrased}
Below, we include the prompt used to generate the paraphrased instances:\\
\begin{lstlisting}
Your task is to restructure the given sentence so that the word to fill in (_, the blank) is now the last word in your restructured sentence that was also a word in the original sentence.
This means, you can add words after the blank, but only if they were not in the original sentence. This is so that the sentence doesn't inadvertently imply a specific answer (e.g., by making one option grammatically or contextually more likely than the other).
Do NOT fill in the _.
In no case should you change anything about the meaning the sentence is conveying, that is, do not add new content to the story that was not in the spirit of the original story.
You can only have one blank in the sentence.
When easily possible, make the sentence sound fluent, while abide by the rules above.

Example 1:
Input: I wanted to build a bathroom on the third floor of the house but I couldn't because the _ would be too full.
Possible tokens to fill in (just for reference): bathroom, floor
Output: I wanted to build a bathroom on the third floor of the house but because it would be too full, that _, I couldn't.
(Notice that this one violates the rules of the last word, but "I couldn't" is a vital part of the story that determines whether _ should be "bathroom" or "floor".)

Example 2:
Input: Jill was on a budget so she only bought a new dress for the ceremony and wore an old hat. She figured the _ would be less noticeable.
Possible tokens to fill in: dress, hat
Output: Jill was on a budget so she only bought a new dress for the ceremony and wore an old hat, figuring that the more noticeable item would be the _.

Example 3:
Input: To make frosting I needed pudding that was at a store 15 minutes away but pre-made frosting was at a store 5 minutes away.  The _ was closer.
Possible tokens to fill in: pudding, frosting
Output: To make frosting I needed pudding that was at a store 15 minutes away but pre-made frosting was at a store 5 minutes away, so the closer choice was the _.

Example 4:
Input: The home that my parents had when I was in school was a lot nicer than my house now because the _ was sophisticated.
Possible tokens to fill in: home, house
Output: The home that my parents had when I was in school was a lot nicer than my house now because of how sophisticated the _ was.

Your task: 
Input: [sentence]
Some possibilities that can replace the token are "[option 1]", or "[option 2]", but either should be able to fill the blank (grammatically speaking).

Reason about how to make this happen, then after thinking, only give the restructured sentence.
Draw inspiration from all of the examples above. What worked previously are eg. cleft sentences, passive voice, relative clauses, appositives, inversions, prepositional phrases, etc.
\end{lstlisting}

The distribution of which model's results are used can be found in Table \ref{paraphrase-models}.

\begin{table}[h]
  \centering
  \begin{tabular}{ll}
    \hline
    \textbf{Model} & \textbf{Freq} \\
    \hline
    GPT-4o & 336 \\
    OpenAI o1-preview & 76 \\
    Gemini 2.0 Flash Thinking Experimental & 105 \\
    Deepseek R1 & 109 \\
    LlaMA 3.2 90B Vision & 95 \\
    Manual & 433 \\
    Original (unchanged) & 31 \\
    \hline
  \end{tabular}
  \caption{\label{paraphrase-models} Number of times each model's output was chosen in the paraphrasing process of the 1,185 retained instances. The bottom two lines contain the number of times a manual adjustment was necessary, and the number of times the original sentence was already in the required paraphrased format.}
\end{table}

\section{Prompt used to categorize sentences} \label{app:promptcategorization}
The first prompt is used to generate reasoning steps to solve the task. As in-context examples, we use instances from the Winograd Schema Challenge. For this step, we use the OpenAI API to prompt gpt-4o-mini-2024-07-18.
\begin{lstlisting}
You are a helpful assistant. Read the instructions carefully.

**INSTRUCTIONS**
Read the Input Text. The Input Text is a text from the WinoGrande benchmark. You get the text, and the two possible options to fill in the _ in the text. 
Think long and hard, and identify the reasoning steps you need to make to decide which option is the correct answer of the Input Text of the TASK.
Provide the reasoning steps concisely. Then, return the correct option.

*IMPORTANT:*
- Your response **must** be in JSON format with the following structure:
  {
      "reasoning": "Your detailed reasoning here.",
      "output": "the correct option to fill in the blank, chosen between Option1 and Option2"
  }
- Do NOT include any additional text outside the JSON object.
- Ensure that the JSON kes are exactly "reasoning" and "output".
- Make sure your reasoning and output relate to the Input Text of the TASK.


**EXAMPLES**
Example Text 1: "The trophy doesn't fit into the brown suitcase because _ is too large. Option 1: The trophy. Option 2: the suitcase."
Example Reasoning 1 : "The object has to be smaller than the container in order to fit inside of it. If the trophy is too large, it does not fit in the suitcase."
Example Output 1 : "The trophy"

Example Text 2: "Joan made sure to thank Susan for all the help _ had recieved. Option 1: Joan. Option 2: Susan."
Example Reasoning 2 : "In social settings, you thank the person that gave you help. Since Joan received the help from Susan, Joan thanked Susan for the help that Joan received."
Example Output 2 : "Joan"

Example Text 3: "The large ball crashed right through the table because _ was made of steel. Option 1: The large ball. Option 2: the table."
Example Reasoning 3 : "We know the ball is large. A large ball made of steel, which is heavy, is more likable to crash through a table."
Example Output 3 : "The large ball"

**TASK**
Input Text: "INPUT_TEXT. Option 1: OPTION1. Option 2: OPTION2."
\end{lstlisting}

The second prompt is given the input text, and the generated reasoning steps from the previous step, to label the instances of one of the five common sense categories. For this step, we use the OpenAI API to prompt gpt-4o-2024-08-06.
\begin{lstlisting}
You are a helpful assistant. Read the instructions carefully.

** INSTRUCTIONS**
Your task is to decide which common sense knowledge categories are present in a text. In the Input Text, you get an example from the WinoGrande benchmark, and the two possible options to fill in the _ in the text. Then, you get the reasoning steps that specify the thought processes.
Read the Input Text and Reasoning Steps carefully, and select one or more common sense knowledge categories in which the Reasoning Steps fit. In other words, which knowledge types are used in the Reasoning Steps?
You can only use categories that are part of the list below. Return the index of the relevant category, following the example below. If multiple categories apply, list all relevant indices separated by commas.
Output only the indices without any additional text or explanations.

**COMMON SENSE CATEGORIES**
1. Physical: Pertains to physical attributes and properties of objects that are relevant to solve the task.
    Examples:
            "The apple is red."
            "The bottle is empty."

2. Social: Involves social norms, roles, and interactions you need to understand to solve the task.
    Examples:
            "She greeted her neighbor."
            "They followed the protocol."

3. Numerical: Relates to numbers and quantities; differences in number or quantity between entities.
    Examples:
            "There are many books on the shelf."
            "He ran 10 miles."

4. Temporal: Concerns time, temporal relations, and eventualities related to important entities of the task (important: NOT about temperature).
    Examples:
            "She arrived before noon."
            "They will meet tomorrow."

5. Spatial: Involves spatial relations (e.g., higher - lower), locations (e.g., north - south), or positions (e.g., behind - in front) that are important to understand to solve the task.
    Examples:
            "The cat is under the table."
            "He walked into the room."


**EXAMPLES**
    Example Reasoning 1 : "The object has to be smaller than the container in order to fit inside of it. If the trophy is too large, it does not fit in the suitcase."
    Relevant Categories 1 : Numerical (3), Spatial (5)
    Output 1: 3, 5

    Example Reasoning 2 : "In social settings, you thank the person that gave you help. Since Joan received the help from Susan, Joan thanked Susan for the help that Joan received."
    Relevant Categories 2: Social (2), Temporal (4)
    Output 2: 2, 4

    Example Reasoning 3 : "We know the ball is large. A large ball made of steel, which is heavy, is more likable to crash through a table."
    Relevant Categories 3: Physical (1), Numerical (3)
    Output: 1, 3

Input Text: "INPUT_TEXT. Option 1: OPTION1. Option 2: OPTION2. Reasoning: REASONING"
\end{lstlisting}

\section{Detailed statistics on memorization checking} \label{app:memorization}
\subsection{Counting the number of contaminated instances}\label{app:contaminationprocedure}
From each of the 1,267 WinoGrande validation instances, we extract the longest $n$-gram that appears at least once in the corpus (i.e. The Pile or RedPajama v1) using the infini-gram API \citep{Liu2024InfiniGram}. For each instance, we find all occurrences of this $n$-gram in documents and extract 100-grams centered on it, ignoring instances with over 100 occurrences. We then prompt \mbox{Open}AI's o1 to verify if the full sentence appears in any of these extracted 100-grams. The prompt can be found in Appendix \ref{app:verifyngramsprompt}. This method allows us to handle inserted characters like LaTeX line breaks, functioning similarly to $k$-skip $n$-grams, though infini-gram doesn't offer the latter capability. This also ensures that contamination can be found even if there are subtle differences in text segments, a criticism that does apply to more naive $n$-gram overlap \citep{DBLP:journals/corr/abs-2406-04244}

The output is positive if the entire instance of the validation set is found in the pre-training dataset.

\subsection{Categorization of contaminated instances}
In Table \ref{tab:contaminated-vs-winogrande}, the category distribution of the contaminated instances of RedPajama v1 is shown. The category distribution of the WinoGrande validation set can be found in Figure \ref{nr_instances_categories}. A two-sample Kolmogorov-Smirnov test on the distributions rejects the zero hypothesis that the distribution of the leaked instances and the true category distribution are drawn from the same underlying distribution ($p = 0.79\%$). Hence, the contaminated instances LLaMA-1 encountered during pre-training are a skewed representation of the true distribution of the WinoGrande validation set.

\begin{table*}[ht!]
\centering
\begin{tabular}{l|c}
\hline
\textbf{Category} & \textbf{RedPajama v1} \\
\hline
\textbf{Social}    & 11  \\
\textbf{Physical}  & 9    \\
\textbf{Spatial}   & 2    \\
\textbf{Numerical} & 4   \\
\textbf{Temporal}  & 1    \\
\hline
\end{tabular}
\caption{The distribution of the contaminated instances in RedPajama v1 according to their categories.}
\label{tab:contaminated-vs-winogrande}
\end{table*}

\subsection{Statistical analysis of contaminated instances}
We analyze whether contamination in the WinoGrande validation set affects model performance using two statistical approaches: (1) comparing logprob differences between truly correct and incorrect answers, and (2) examining binary classification rates. The classification rate represents the proportion of correct predictions made by the model: a value of 1 means the model correctly identified the answer, while 0 indicates an incorrect prediction. For both approaches, we test contaminated instances from RedPajama v1 against non-contaminated instances across Llama-1 models using one-sided tests (Mann-Whitney U for logprobs and Fisher's exact for classification rates). We formulate the following hypotheses:
\begin{itemize}
    \item $H_0$: There is no difference in performance (logprob differences/classification accuracy) between contaminated and non-contaminated instances.
    \item $H_a$: Performance is greater for contaminated instances than for non-contaminated instances.
\end{itemize}

The p-values for both tests can be found in Table \ref{tab:statistical-tests}. Neither test showed statistical significance, and thus the null hypotheses cannot be rejected.

\begin{table*}[ht!]
\centering
\begin{tabular}{l|c c}
\hline
\textbf{Model} & \textbf{Mann-Whitney U $\mathbf{p}$-value} & \textbf{Fisher's exact $\mathbf{p}$-value} \\
\hline
\textbf{Llama 7B}       & 0.0539 & 0.054 \\
\textbf{Llama 13B}      & 0.2665 & 0.267 \\
\textbf{Llama 30B}      & 0.0945 & 0.095 \\
\textbf{Llama 65B}      & 0.2573 & 0.257 \\
\hline
\end{tabular}
\caption{Statistical test p-values checking the effect of contamination in the WinoGrande validation set (RedPajama v1) on Llama 1 models' performance, measured by logprob differences (Mann-Whitney U) and binary classification rates (Fisher's exact).}
\label{tab:statistical-tests}
\end{table*}


\subsection{Correlation between $n$-gram length/count and performance on contaminated instances}
We conducted two visual correlation analyses for all instances in the WinoGrande validation set against the logit difference between ground truth correct and incorrect answers. The first analysis examined the correlation with the length of the longest $n$-gram sequence that appears at least once in the pre-training data, as detailed in Appendix \ref{app:contaminationprocedure}. The second analysis focused on $n$-gram frequency, measuring how often the longest $n$-gram occurs in the pre-training data. Figure 1 displays scatter plots for both analyses, and neither reveals any visible correlation.

\begin{figure}[htbp]
    \centering
    \subfigure[Correlation with n-gram length]{%
        \includegraphics[width=0.45\textwidth]{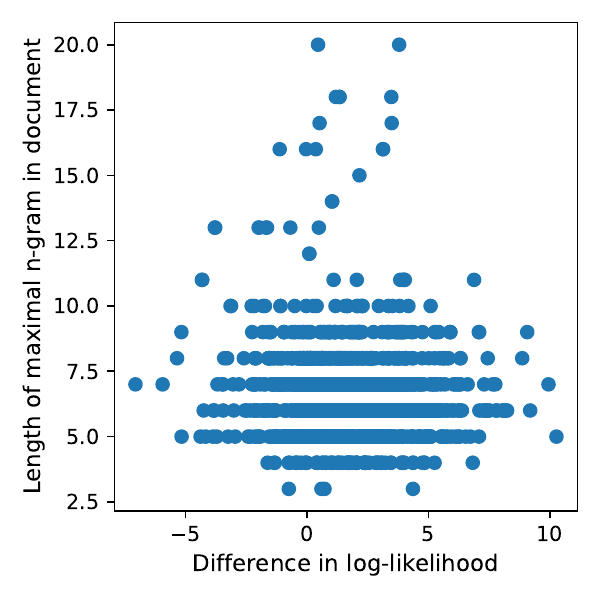}%
        \label{fig:ngram_length}%
    }\hfill
    \subfigure[Correlation with n-gram count]{%
        \includegraphics[width=0.45\textwidth]{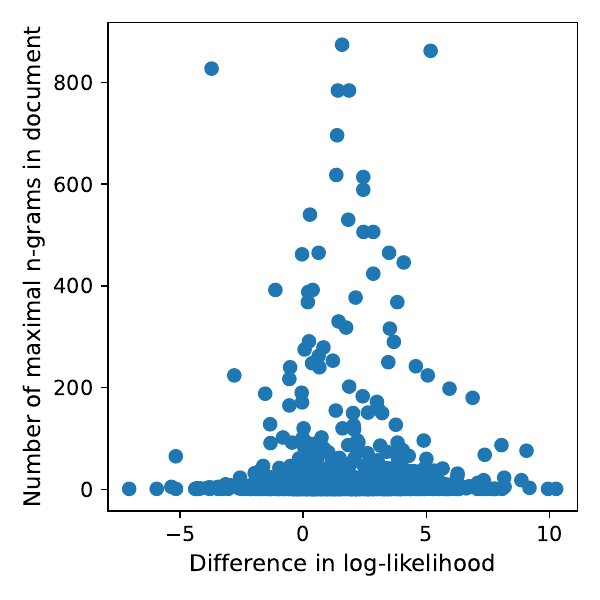}%
        \label{fig:ngram_count}%
    }
    \caption{Scatter plots showing correlation for all instances in the WinoGrande validation set against the logit difference between ground truth correct and incorrect answers. (a) Correlation with n-gram length. (b) Correlation with n-gram count.}
    \label{fig:correlation_analyses}
\end{figure}

\section{Prompt used to verify $n$-grams in a pre-training dataset}\label{app:verifyngramsprompt}
\begin{lstlisting}
Given is a sentence. Below that is an n-gram that occurs in the sentence. Below that are some numbered documents, with the number between parentheses, that all contain the n-gram.

Does at least one of the documents also contain the entire sentence, regardless of the occurrence of the the n-gram? If no, respond only "no". If yes, give one document number and respond only "yes: <document number>".
Do not respond anything else.

Sentence: "[sentence]"
N-gram: "[n-gram]"
Documents:
```
(1) [document 1 excerpt]
(2) [document 2 excerpt]
...
```
\end{lstlisting}

\end{document}